\documentclass[10pt,twocolumn,letterpaper]{article}

\usepackage[pagenumbers]{cvpr} 
\usepackage{tabularx}
\usepackage{multirow}
\usepackage[utf8]{inputenc}
\usepackage[T1]{fontenc}
\usepackage{booktabs}
\usepackage{bbm}
\usepackage{pifont}
\usepackage{placeins}
\usepackage{capt-of} 
\newcommand{\cmark}{\ding{51}}
\newcommand{\xmark}{\ding{55}}

\definecolor{cvprblue}{rgb}{0.21,0.49,0.74}
\usepackage[pagebackref,breaklinks,colorlinks,allcolors=cvprblue]{hyperref}

\title{RIGOR: Rig-Informed Geometry for Omnidirectional Reconstruction}

\author{Tingjun Huang\\
ETH Zürich\\
{\tt\small huangti@ethz.ch}
\and
Dmitry Rudshin\\
ETH Zürich\\
{\tt\small drudshin@ethz.ch}
\and
Mathieu Meyer\\
ETH Zürich\\
{\tt\small mathmeyer@ethz.ch}
\and
Pietro Bonazzi\\
ETH Zürich\\
{\tt\small pbonazzi@ethz.ch}
\and
Marc Pollefeys\\
ETH Zürich\\
{\tt\small marc.pollefeys@inf.ethz.ch}
\and
Emilia Szymańska\\
Hilti AG\\
{\tt\small emilia.szymanska@hilti.com}
}

\usepackage{xcolor}
\usepackage{colortbl}
\usepackage{soul}

\definecolor{lightgreen}{RGB}{200,255,200}
\definecolor{lightorange}{RGB}{252,210,153}
\definecolor{lightyellow}{RGB}{255,255,180}
\definecolor{lightpink}{RGB}{255,182,193}

\begin{document}
\twocolumn[{%
\maketitle
\vspace{-0.45cm}
\begin{center}
    \includegraphics[width=\textwidth]{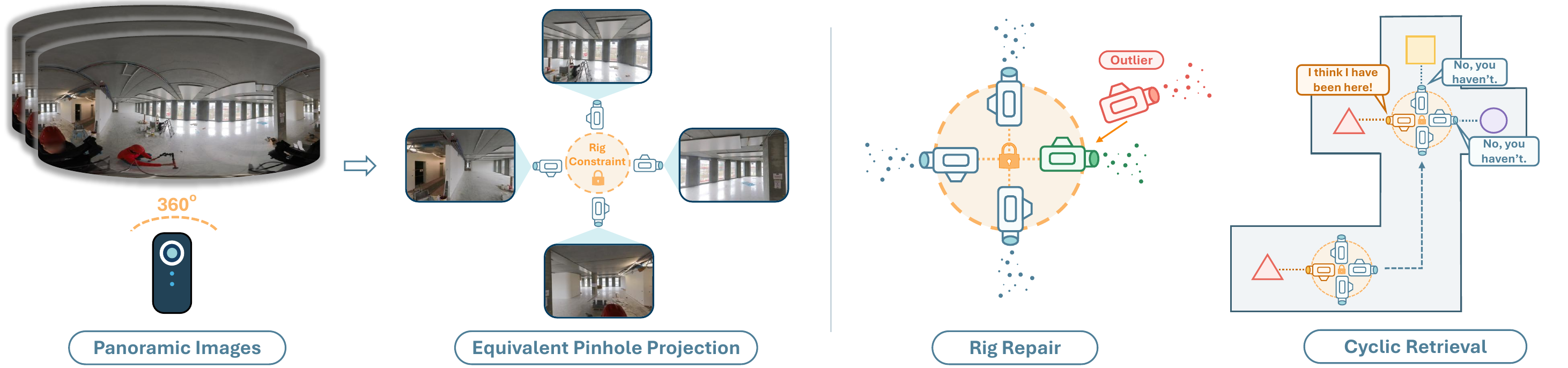}
    \captionof{figure}{Conceptual illustration of the RIGOR principle, providing consistency constraints for local and global reconstruction.}
    \label{fig:teaser}
\end{center}
\vspace{0.05cm}
}]

\begin{abstract}
    Recent developments in feed-forward 3D reconstruction resulted in models which can recover dense scene representations and camera motion solely from an image stream. However, such predictions are prone to becoming inconsistent over long trajectories, specifically in demanding environments with repetitive structures, weak textures and dynamic objects or people. One way to mitigate those challenges is to use an omnidirectional camera, which provides wide spatial coverage and captures richer visual information. Yet, the majority of models do not offer support for 360-degree imagery or require additional fine-tuning. To bridge these two aspects, we present RIGOR: a large-scale reconstruction pipeline for gravity-aligned omnidirectional videos that retains a frozen feed-forward perspective backbone and exploits each panorama as a four-view virtual rig. The rig structure is used to detect and repair locally inconsistent predictions, to retrieve loop closures through cyclic four-view consensus, and to geometrically verify candidate revisits before global optimization. Verified constraints drive a Sim(3) pose graph that corrects accumulated rotation, translation, and scale drift along the sequence. We demonstrate that the proposed consistency mechanisms improve both trajectory accuracy and reconstructed geometry over a feed-forward baseline on challenging construction-site sequences. The code is made available under this link: \href{https://github.com/TangentH/RIGOR}{https://github.com/TangentH/RIGOR}.
\end{abstract}

\section{Introduction}

Dense three-dimensional reconstruction methods offer an attractive map representation for applications spanning localization, robotic navigation, semantics-driven tasks, and digital twins. Traditionally, numerous methods in Simultaneous Localization and Mapping (SLAM) and Structure from Motion (SfM) have been developed to produce such maps~\cite{picard2023surveyrealtime3dscene, survey3d}, but they typically rely on sufficient viewpoint diversity, reliable image correspondences, and geometric optimization to recover scene structure and camera motion. With recent advances in computer vision, it has become possible to reconstruct scene geometry from only a small number of uncalibrated images without requiring known camera poses~\cite{vggt, dust3r, da3}. These learned approaches, commonly referred to as feed-forward 3D reconstruction methods, have become an active research area due to their simple inference pipelines and strong out-of-the-box performance.

A particularly interesting type of imagery to be used in such networks is panoramic imagery captured by omnidirectional cameras~\cite{survey360}. These cameras, becoming cheaper and more widely available, offer an instantaneous capture of the full $360^\circ$ field of view, which makes them particularly beneficial in environments poor in visual features and with dynamic content that could otherwise partially or fully occlude a narrower field of view. However, perspective-image datasets are still dominant, and the majority of feed-forward reconstruction networks are designed for perspective inputs, therefore generalizing poorly to panoramic inference. The few existing methods targeting panoramic imagery~\cite{panovggt,vggt360} primarily focus on reconstruction or geometric consistency within individual panoramas or short image sets, rather than long-trajectory reconstruction and global consistency over extended sequences.

To address this gap, we propose a Rig-Informed Geometry for Omnidirectional Reconstruction (RIGOR) mechanism that improves both local and global trajectory and geometric consistency in panoramic feed-forward 3D reconstruction. We assume gravity-aligned panoramas, which can be obtained from the inertial sensing available in typical $360^\circ$ cameras and is treated only as preprocessing. We project each panorama into four perspective views and treat them as a virtual camera rig. Since these views share the same optical center and have known relative orientations, the rig provides an explicit geometric consistency prior that can be exploited without modifying or fine-tuning the underlying feed-forward reconstruction model. We use this structure to improve local prediction consistency, perform rig-aware loop closure across revisited locations, thereby enabling globally consistent reconstruction over long trajectories, as illustrated in Fig.~\ref{fig:teaser}.

In summary, in this paper we make the following contributions:
\begin{itemize}
    \item We formulate RIGOR, a virtual-rig-based method that exploits the known co-location and relative orientations of perspective views as constraints for detecting and repairing inconsistent pose and geometry predictions;
    \item We provide a rig-aware capture representation for loop retrieval that cyclically aligns descriptors from all virtual views, followed by a joint geometric verification of the retrieved candidates;
    \item We demonstrate the performance of the pipeline on the construction site captures from the Hilti-Trimble-Oxford dataset\cite{hiltichallenge} with respect to the consistency of both the trajectory and geometry.
\end{itemize}

\section{Related Work}

We organize the relevant prior work into three groups, summarized together with our method in Tab.~\ref{tab:related_work}.

\paragraph{Feed-forward multi-view reconstruction.}
Feed-forward 3D reconstruction methods aim to infer scene geometry and
camera information directly from image collections, reducing reliance on
conventional multi-stage reconstruction pipelines. DUSt3R~\cite{dust3r}
introduced a general formulation that predicts pairwise pointmaps from
uncalibrated image pairs, while MASt3R~\cite{mast3r} extends DUSt3R with
dense local features and more accurate, efficient image matching.
MUSt3R~\cite{must3r} extends this paradigm from pairs to multiple views,
using a memory mechanism to scale reconstruction to large image
collections. VGGT~\cite{vggt} jointly predicts camera parameters, depth, point maps, and point tracks,
while Depth Anything 3~\cite{da3} predicts depth and camera geometry from
an arbitrary number of visual inputs. MapAnything~\cite{mapanything}
extends feed-forward reconstruction to heterogeneous geometric inputs
and metric reconstruction, while $\pi^3$~\cite{pi3} removes the
fixed-reference-view dependency through permutation-equivariant visual
geometry prediction. However, these general-purpose methods are not specifically designed for panoramic input, and most of them lack explicit mechanisms for revisit detection and corrective optimization in long sequences. 
\paragraph{Persistent reconstruction and learned SLAM.}
For temporally ordered observations, a related line of work extends learned geometric priors to persistent reconstruction and online mapping. Spann3R~\cite{spann3r} introduces a spatial memory that enables incremental reconstruction by predicting pointmaps in a common coordinate system, avoiding test-time global alignment. CUT3R~\cite{cut3r} further adopts a recurrent state that is continuously updated with incoming observations, producing metric-scale pointmaps in a common coordinate system and supporting both static and dynamic scenes. SLAM3R~\cite{slam3r} instead combines feed-forward local reconstruction with incremental registration of local pointmaps into a globally consistent scene. Learned SLAM systems build on related geometric priors while explicitly maintaining camera trajectories and maps: MASt3R-SLAM~\cite{mast3rslam} combines MASt3R pointmaps with tracking, local map fusion, loop closure, and global optimization, while VGGT-SLAM 2.0~\cite{vggtslam2} incrementally aligns VGGT submaps using a keyframe-level factor graph and verifies loop candidates using VGGT attention. MegaSaM~\cite{megasam} targets unconstrained monocular videos, combining a differentiable SLAM framework with monocular depth priors and motion probability maps for camera and depth estimation in dynamic scenes. 

\begin{figure*}[t]
 \centering
 \includegraphics[width=\textwidth]{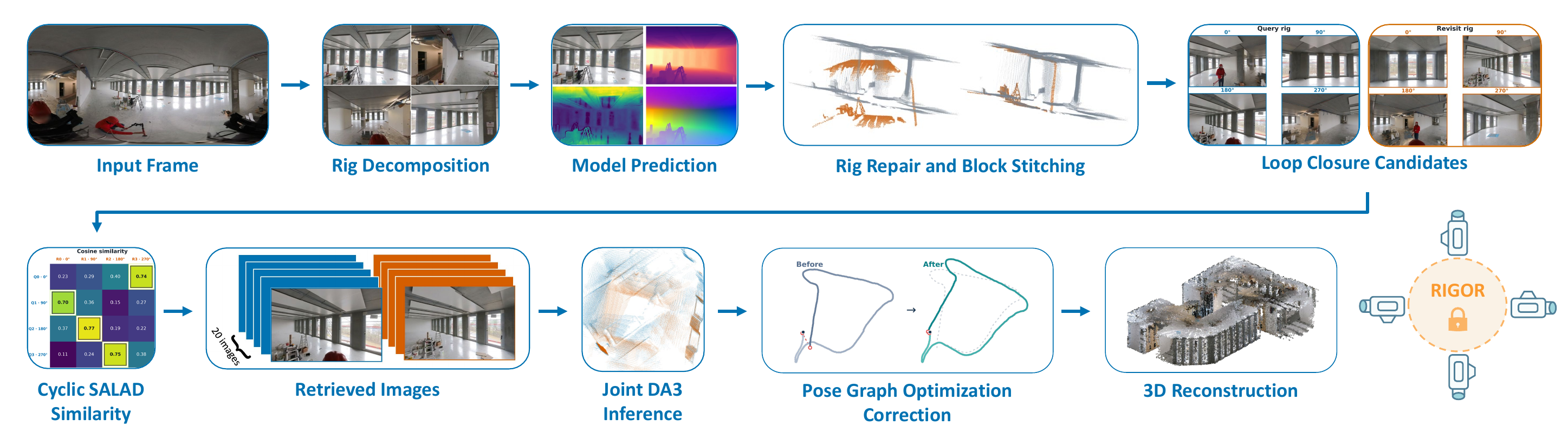}
\caption{RIGOR retains a frozen perspective predictor and uses the known
four-view panorama rig for validated local repair and cyclic multi-view loop
retrieval. Joint feed-forward prediction verifies the retrieved candidates
before accepted loop measurements enter a global $\mathrm{Sim}(3)$ pose graph
for reconstruction fusion.}
 \label{fig:pipeline}
\end{figure*}

\paragraph{Panoramic reconstruction and SLAM.} Panoramic reconstruction introduces additional challenges due to the geometric distortions and spherical structure of equirectangular imagery. PanoVGGT~\cite{panovggt} addresses panoramic 3D reconstruction directly from equirectangular inputs, using spherical-aware representations to predict camera poses, dense depth, and geometrically consistent scene structure. VGGT-360~\cite{vggt360} instead follows a training-free strategy that projects an equirectangular panorama into perspective views, performs multi-view 3D reasoning with VGGT-like models, and reprojects the resulting 3D predictions back to the panorama. For panoramic visual SLAM and odometry, OpenVSLAM~\cite{openvslam} supports perspective, fisheye, and equirectangular cameras within a feature-based SLAM framework with explicit loop detection and global optimization, while PatchMatch-Stereo-Panorama~\cite{patchmatch} builds on OpenVSLAM and adds panorama-aware PatchMatch stereo for dense reconstruction. OmniDSO~\cite{omnids0} extends direct sparse odometry to omnidirectional cameras using a unified camera model, while PanoAir~\cite{panoair} combines equirectangular imagery with IMU measurements and panoramic loop closure for visual-inertial SLAM. Our method similarly targets temporally ordered panoramic observations, but combines a frozen perspective feed-forward model with four-view geometric consistency, rig-aware loop retrieval and verification, and global loop-based correction (Fig.~\ref{fig:pipeline}).

\begin{table}[t]
    \centering
\caption{
Comparison of representative reconstruction systems, categorized by the
support of panoramic images (\textit{Pano.}), dense reconstruction
(\textit{Dense}), long sequences (\textit{Long seq.}), explicit revisit
detection with corrective optimization (\textit{Loop}), and whether
panorama-specific learned training was required (\textit{Pano train.}).
\mbox{``--''} denotes that the criterion is not applicable.
}
    \label{tab:related_work}

    \setlength{\tabcolsep}{2.5pt}
    \renewcommand{\arraystretch}{1.08}
    \footnotesize

    \begin{tabularx}{\columnwidth}{@{}
        >{\raggedright\arraybackslash}X
        ccccc
    @{}}

        \toprule
        \multicolumn{1}{c}{\begin{tabular}[t]{@{}c@{}}Method\end{tabular}} &
        \begin{tabular}[t]{@{}c@{}}Pano.\end{tabular} &
        \begin{tabular}[t]{@{}c@{}}Dense\end{tabular} &
        \begin{tabular}[t]{@{}c@{}}Long\\seq.\end{tabular} &
        \begin{tabular}[t]{@{}c@{}}Loop\end{tabular} &
        \begin{tabular}[t]{@{}c@{}}Pano\\train.\end{tabular} \\

        \midrule

        \multicolumn{6}{@{}l}{\textit{Feed-forward reconstruction}} \\
        \addlinespace[0.15em]

        DUSt3R, MASt3R,
        & \multirow{2}{*}{\xmark}
        & \multirow{2}{*}{\cmark}
        & \multirow{2}{*}{\xmark}
        & \multirow{2}{*}{\xmark}
        & \multirow{2}{*}{\xmark} \\[-0.25em]
        VGGT, DA3, $\pi^3$, MapAnything
        & & & & & \\

        MUSt3R
        & \xmark & \cmark & \cmark & \xmark & \xmark \\


        \addlinespace[0.35em]
        \midrule
        \multicolumn{6}{@{}l}{\textit{Persistent reconstruction / learned SLAM}} \\
        \addlinespace[0.15em]

        Spann3R, CUT3R, SLAM3R, 
        & \multirow{2}{*}{\xmark}
        & \multirow{2}{*}{\cmark}
        & \multirow{2}{*}{\cmark}
        & \multirow{2}{*}{\xmark}
        & \multirow{2}{*}{\xmark} \\
        [-0.25em]
        MegaSaM
        & & & & & \\

        MASt3R-SLAM, VGGT-SLAM 2.0 & \xmark & \cmark & \cmark & \cmark & \xmark \\

        \addlinespace[0.35em]
        \midrule
        \multicolumn{6}{@{}l}{\textit{Panoramic reconstruction / SLAM}} \\
        \addlinespace[0.15em]

        PanoVGGT
        & \cmark & \cmark & \xmark & \xmark & \cmark \\

        VGGT-360
        & \cmark & \cmark & \xmark & \xmark & \xmark \\

        OpenVSLAM, PanoAir
        & \cmark & \xmark & \cmark & \cmark & -- \\
        
        PatchMatch-Stereo-Panorama & \cmark & \cmark & \cmark & \cmark & -- \\

        OmniDSO
        & \cmark & \xmark & \cmark & \xmark & -- \\

        \addlinespace[0.35em]
        \midrule

        \textbf{Our method}
        & \cmark & \cmark & \cmark & \cmark & \xmark \\

        \bottomrule
    \end{tabularx}
\end{table}

\section{Method}
\label{sec:method}
\subsection{Preprocessing}

Since many consumer-grade $360^\circ$ cameras either natively use an inertial measurement unit (IMU) for leveling or provide IMU measurements for custom leveling procedures, we assume that the processed video is already gravity-aligned. To suppress dynamic foreground geometry, we use Grounded Segment Anything~\cite{ren2024grounded} to mask people. The static footprint of the capture device is removed using a separately calibrated mask projected from the equirectangular image into the four perspective views. The two masks are merged and applied by suppressing post-inference confidence; the RGB inputs to the frozen predictor remain unchanged.

\subsection{Virtual Rig and Feed-Forward Inference}
\label{sec:overview}

The input is a temporally ordered sequence of $T$ gravity-aligned panoramas
$\{E_t\}_{t=1}^{T}$. Each
panorama is projected into four perspective images,
\begin{equation}
 I_t^v=\Pi(E_t;Q_v),\quad Q_v=R_y(\pi v/2),\quad v\in\{0,1,2,3\}
 \label{eq:virtual_views}
\end{equation}
where $I_t^v$ is perspective view $v$ of panorama $t$, $\Pi$ is the
equirectangular-to-perspective projection, and $Q_v$ is its known yaw rotation
relative to the panorama reference orientation. These views form a
\emph{virtual rig}: their optical centers coincide and their relative rotations
are known, but they provide neither stereo parallax nor metric scale. We use
this structure for local prediction repair and capture-level loop retrieval
and geometric verification for closing the loop; the verified measurements are then consumed by a
$\mathrm{Sim}(3)$ graph optimizer.

Depth Anything 3 (DA3)~\cite{da3} predicts camera poses, effective intrinsics, per-pixel depths, and confidences for short overlapping blocks of perspective images. The rendering field of view (FoV) determines each image's angular coverage, whereas the DA3-predicted intrinsics are those associated with its predicted depth and pose and are therefore used for back-projection. 

\subsection{Sequential Reconstruction with Rig Correction}
\label{sec:sequential}

Let block $k$ be a consecutive group of perspective images given to DA3.
Adjacent blocks share images; confidence-weighted pointmap alignment estimates a relative similarity transform
$S_{k,k+1}\in\mathrm{Sim}(3)$ and places their predictions in one reconstruction
frame. This overlap estimates relative scale between blocks, while the full reconstruction remains defined up to one arbitrary global scale. Ownership is deterministic and independent of loop closure: every non-terminal block discards its trailing 16 shared images, the following block owns them, and the terminal block retains its remaining tail. Shared predictions are still used for alignment, but only the owner's points are fused.

For panorama \(t\), the frozen backbone predicts camera centers
\(\widehat C_t^v\) and rotations \(\widehat R_t^v\). Under the known
virtual-rig geometry, these predictions should be close to a shared-center
model,
\begin{equation}
\widehat C_t^v \approx C_t,\qquad
\widehat R_t^v \approx U_t Q_v
\label{eq:rig_constraint}
\end{equation}
where \(C_t\) and \(U_t\) are the fitted common center and base orientation,
respectively, and \(Q_v\) is the known relative orientation of virtual view
\(v\). We first test the complete four-view set and, only if it is invalid, its four three-view subsets. A subset is valid when its center residual is at most $0.05$ times the median positive predicted depth of the panorama and its rotation residual is at most $5^\circ$. Among valid subsets of the current cardinality, we select the fit
with the lowest sum of squared residuals normalized by these bounds; hence the largest valid cardinality is always preferred. A valid four-view fit is left unchanged. A selected three-view fit makes only the remaining view $v^*$ a repair candidate. With fewer than three valid views, the method abstains.

The candidate keeps its DA3-predicted depth $D_t^{v^*}$, confidence, and
intrinsics, but is back-projected using the fitted rig pose:
\begin{equation}
 X_t^{v^*}(p)=C_t+D_t^{v^*}(p)\,U_tQ_{v^*}K_{t,v^*}^{-1}\widetilde p
 \label{eq:rig_reprojection}
\end{equation}
where $X_t^{v^\ast}(p)$ is the repaired 3D point corresponding to pixel
$p$ in the candidate view $v^\ast$, $\widetilde p$ its homogeneous coordinate, and
$K_{t,v^*}$ the effective intrinsics predicted by DA3 for the view $v^*$. Adjacent yaw views have
a small angular overlap because their directions are spaced by $90^\circ$ while the rendered FoV is $95^\circ$. Within this overlap, we form mutual ray
matches and measure their normalized 3D distances before and after repair. The proposal is accepted only with at least 100 matches, a strictly lower median distance, and a 90th percentile that does not increase. Otherwise the original prediction is
retained. Thus a proposal changes only the pose and resulting pointmap of the isolated view.

\subsection{Four-View Loop Closure}
\label{sec:loops}

A frozen SALAD encoder~\cite{salad} produces a normalized descriptor $z_t^v$
for every perspective image. Since views within each panorama are rendered at
fixed $90^\circ$ relative yaw offsets, correspondences between two panoramas
must preserve the same cyclic structure. For two panoramas $t$ and $s$, let
\begin{equation*}
 M_{ts}(v,u)=\langle z_t^v,z_s^u\rangle
\end{equation*}
denote the descriptor similarity between view $v$ of panorama $t$ and view $u$ of panorama $s$, where $\langle\cdot,\cdot\rangle$ is the inner product.
Because the descriptors are normalized, this inner product is cosine similarity.

We jointly score the four possible cyclic assignments:
\begin{equation}
\begin{aligned}
m_q(t,s)
&=\frac{1}{4}\sum_{v=0}^{3}
M_{ts}\!\left(v,(v+q)\bmod 4\right),\\
q^*
&=\arg\max_{q\in\{0,1,2,3\}} m_q(t,s),
\end{aligned}
\label{eq:cyclic_retrieval}
\end{equation}
We enumerate every panorama pair with $|t-s|>80$ and retain it only if
$m_{q^*}(t,s)>0.65$ and all four similarities on the selected diagonal are at least $0.50$. Capture-level non-maximum suppression then discards a weaker pair when both its endpoints lie within 25 panoramas of the corresponding endpoints of a stronger pair. The shared cyclic assignment permits opposite headings but does not let one view propose a loop independently.

Each retained candidate is verified by a joint DA3 prediction. On the selected cyclic diagonal, its highest-scoring entry defines the representative view pair. Around each representative view, we take a block-constrained window of up to
20 consecutive perspective images per side. Concatenating both sides gives one
joint DA3 context of at most 40 images, predicted in a common frame $J$. The same rig fitting and
accepted local repair are applied to these joint predictions. Similarity transforms $H_t$ and $H_s$ attach $J$ to the corresponding stored blocks, giving
\begin{equation}
 Z_{s\leftarrow t}=H_sH_t^{-1}
 \label{eq:loop_measurement}
\end{equation}
where $H_t$ maps $J$ to the block containing panorama $t$, and likewise for $H_s$. Each attachment residual must satisfy a fixed bound, and their induced relative scale must lie in $[0.8,1.25]$. On each side, at least $80\%$ of eligible complete panoramas must admit a three- or four-view rig fit. Finally, confidence-filtered point sets from the two sides must have sufficient
bidirectional coverage and mutual-nearest-neighbour agreement in $J$. No cross-side ICP is used: joint prediction supplies the shared frame, whereas
registration could force distinct but visually similar corridors to agree.

\subsection{Sim(3) Pose Graph and Fusion}
\label{sec:graph}

Sequential block alignments and verified loop measurements form a pose graph
$\mathcal G=(\mathcal V,\mathcal E)$, where $\mathcal V$ indexes the blocks and
$\mathcal E$ contains the sequential and verified loop edges.
Let $G_k\in\mathrm{Sim}(3)$ map block $k$ into the global reconstruction
frame. Because $Z_{j\leftarrow i}$ maps block $i$ into block $j$, a consistent
edge $(i,j)\in\mathcal E$ satisfies
$Z_{j\leftarrow i}G_i^{-1}G_j=I$. We optimize
\begin{equation}
\begin{aligned}
r_{ij}(G_i,G_j)
&=\operatorname{Log}\!\left(Z_{j\leftarrow i}G_i^{-1}G_j\right),\\
\{G_k^\star\}
&=\arg\min_{\{G_k\}}\sum_{(i,j)\in\mathcal E}
\left\lVert r_{ij}(G_i,G_j)\right\rVert_2^2,
\qquad G_0=I.
\end{aligned}
\label{eq:sim3_graph}
\end{equation}

This gauge choice selects a representative of the arbitrary global
similarity gauge. Node updates may change rotation, translation, and
log-scale, allowing the graph to redistribute scale drift along the
trajectory. Optimized transforms and accepted rig corrections are applied
consistently to camera poses and pointmaps before owner-point fusion.

\paragraph{Implementation details.}
We render four $95^\circ$ perspective views per panorama at fixed
$90^\circ$ yaw intervals. DA3 processes $N=32$
perspective images per sequential block with overlap $O=16$. Dynamic/device
masks leave DA3's RGB input unchanged and suppress the corresponding
post-inference confidences before alignment and fusion. After owner-point
fusion, a fixed 5cm occupancy-support cleanup removes isolated voxels. Joint
DA3 inference is performed only for candidates retained by cyclic retrieval.
Thresholds were selected during development and frozen before the final
benchmark-wide rerun; one parameter set is used for every sequence. 
Additional implementation details are provided in the supplementary material.

\section{Experiments}
\label{sec:experiments}

\subsection{Experimental Protocol}

In this section, we describe the evaluation metrics selected for comparing the methods: those related to the predicted camera poses and those related to the reconstructed geometry. Additionally, we present the registration procedure used to align the predicted point clouds with the ground-truth point clouds, and discuss the selection of the benchmark and baselines.

\paragraph{Benchmark.}
To evaluate the method, a suitable dataset had to be selected. Our main focus was to assess the applicability of the proposed approach in demanding scenarios, particularly in scenes with few visual features and repetitive structures. The recent Hilti--Trimble--Oxford dataset~\cite{hiltichallenge} provides 30 panoramic sequences captured in construction environments, which were intentionally designed to be challenging for the associated competition. Ground-truth trajectories are available for all sequences, and, upon request, we additionally obtained dense LiDAR point clouds for 29 of them. The evaluation is therefore conducted on 30 trajectories and 29 point clouds.

\paragraph{Trajectory evaluation metrics.} 

Let the estimated and ground-truth camera positions at timestamp \(t\) be \(\mathbf{p}_t\in\mathbb{R}^3\) and \(\mathbf{p}_t^{\mathrm{gt}}\in\mathbb{R}^3\), respectively. Before evaluation, we estimate a rigid alignment
\((\mathbf{R},\mathbf{t})\in\mathrm{SE}(3)\) between the predicted and ground-truth trajectories; importantly, no scale correction is applied. We report the Absolute Trajectory Error (ATE) as the root-mean-square Euclidean position error
$$
\mathrm{ATE} = \sqrt{ \frac{1}{N} \sum_{t=1}^{N} \left\| \mathbf{R}\mathbf{p}_t+\mathbf{t} - \mathbf{p}_t^{\mathrm{gt}}\right\|_2^2
},
$$

and a position-only translational Relative Pose Error over a
$10\,\mathrm{s}$ horizon. Let
$\widetilde{\mathbf p}_i=\mathbf R\mathbf p_i+\mathbf t$ be the rigidly aligned
estimated position. For each timestamp $t_i$, we choose the closest future
timestamp $t_{j(i)}$ satisfying
$|(t_{j(i)}-t_i)-10\,\mathrm{s}|\leq0.05\,\mathrm{s}$. We report
\begin{equation}
\mathrm{RPE}_{10\mathrm{s}}=
\sqrt{\frac{1}{|\mathcal P_{10}|}
\sum_{i\in\mathcal P_{10}}
\left\|
(\widetilde{\mathbf p}_{j(i)}-\widetilde{\mathbf p}_i)-
(\mathbf p^{\mathrm{gt}}_{j(i)}-\mathbf p^{\mathrm{gt}}_i)
\right\|_2^2}.
\end{equation}
Ground-truth positions are associated by timestamp, with interpolation allowed
only across gaps no larger than $0.75\,\mathrm{s}$. Every reported trajectory
passes the $99\%$ association-coverage requirement.

\paragraph{Geometry evaluation metrics.}
Let \(\mathcal{X}\) and \(\mathcal{Y}\) denote the reconstructed and ground-truth point sets restricted to the evaluation Region of Interest (ROI). Each reconstructed point cloud is aligned independently to the ground truth for evaluation using the procedure described in the following sections. We report the symmetric Chamfer-\(L_1\) distance (denoted as $CD$)

$$ d(\mathcal{X},\mathcal{Y}) = \frac{1}{|\mathcal{X}|} \sum_{\mathbf{x}\in\mathcal{X}} \min_{\mathbf{y}\in\mathcal{Y}} \|\mathbf{x}-\mathbf{y}\|_2, $$
$$\mathrm{CD}(\mathcal{X},\mathcal{Y}) = \frac{1}{2} \left( d(\mathcal{X},\mathcal{Y}) + d(\mathcal{Y},\mathcal{X}) \right). $$

as well as the \(F\)-score at distance threshold
\(\tau=0.25\,\mathrm{m}\). Defining precision $P_\tau$ and recall $R_\tau$ as

$$
P_\tau
= \frac{1}{|\mathcal{X}|} \sum_{\mathbf{x}\in\mathcal{X}} \mathbbm{1} \!\left[ \min_{\mathbf{y}\in\mathcal{Y}} \|\mathbf{x}-\mathbf{y}\|_2 < \tau \right], 
$$
$$ R_\tau = \frac{1}{|\mathcal{Y}|} \sum_{\mathbf{y}\in\mathcal{Y}} \mathbbm{1} \!\left[ \min_{\mathbf{x}\in\mathcal{X}} \|\mathbf{y}-\mathbf{x}\|_2 < \tau \right], 
$$

the reported score is

$$
F_\tau=\frac{2P_\tau R_\tau}{P_\tau+R_\tau}, \qquad \tau=0.25\,\mathrm{m}.
$$

\paragraph{ROI and Registration.}
    Ground-truth (GT) point clouds are obtained from a LiDAR sensor and contain substantial background geometry, e.g., trees or streets outside the navigated building, as well as mirror-like artifacts caused by glass surfaces. To ensure that the evaluation focuses solely on the geometric correctness of the relevant scene, we define a Region of Interest (ROI) over which all metrics are computed. Since we are interested in reconstructing the indoor scene, we take advantage of the fact that the Hilti-Trimble-Oxford dataset provides ground-truth trajectories, and consequently LiDAR point clouds, registered with respect to the floor plans. Each floor plan is represented as a binary mask indicating walls and windows, which effectively defines the building boundary. We therefore discard LiDAR points that fall outside the building footprint in the $XY$ plane, with a 5 cm margin. Additionally, since the point cloud is gravity-aligned, we identify peaks in the histogram of point positions along the $Z$-axis, which are expected to correspond to the floor and ceiling. We then remove points above the estimated ceiling level and below the floor plane. The resulting prism, bounded laterally by the floor-plan interior and vertically by the estimated floor and ceiling, defines the ROI.
    
    To evaluate the geometric correctness of the reconstructed environments, the predicted and ground-truth point clouds must be registered in a common coordinate system and aligned. Each reconstructed point cloud is independently registered using a GT-assisted, evaluation-only $\mathrm{Sim}(3)$ initializer estimated from the predicted and reference trajectories, followed by scale-adjusting ICP against the ROI LiDAR cloud. The ICP-based alignment is retained only if it improves the registration objective; otherwise, the trajectory-based initialization is used. This oracle alignment protocol evaluates reconstructed shape after global similarity alignment; it does not assess recovery of metric scale or absolute map placement.

\paragraph{Compared systems.}
We initially selected PanoVGGT, VGGT-SLAM2, MASt3R-SLAM, PatchMatch (also known as Stella Dense), and VGGT as representative state-of-the-art baselines covering different classes of methods. Upon closer inspection, however, we found that most of these methods fail on a substantial fraction of the considered sequences, often producing severely incomplete or geometrically inconsistent reconstructions. Since such failure cases are more informative when inspected visually than summarized by aggregate geometric metrics, we report these methods primarily in the qualitative results section. For the quantitative evaluation, we therefore focus on the strongest and most relevant baselines that consistently produce reconstructions suitable for meaningful geometric comparison.

\emph{DA3-Sequential} is the sequential DA3 reconstruction using the same four-view inputs without a non-local graph update. \emph{DA3-Legacy} uses the same frozen DA3 checkpoint and front-end inputs, but retains the original single-view retrieval, verification, and graph protocol; it therefore serves as a same-backbone predecessor with cyclic capture retrieval, joint geometric verification, and virtual-rig repair disabled. \emph{PanoVGGT} is extended to long sequences with the same overlapping-block stitching as DA3 and is referred to as \emph{PanoVGGT} hereafter; it is the best-performing baseline and is therefore retained for quantitative comparison. \emph{Ours w/o rig repair} is used only for the controlled repair study, whereas \emph{Ours} denotes the complete RIGOR method.

\subsection{Results}

\paragraph{Quantitative results.}

\begin{figure}[t]
     \centering
     \includegraphics[width=\columnwidth]{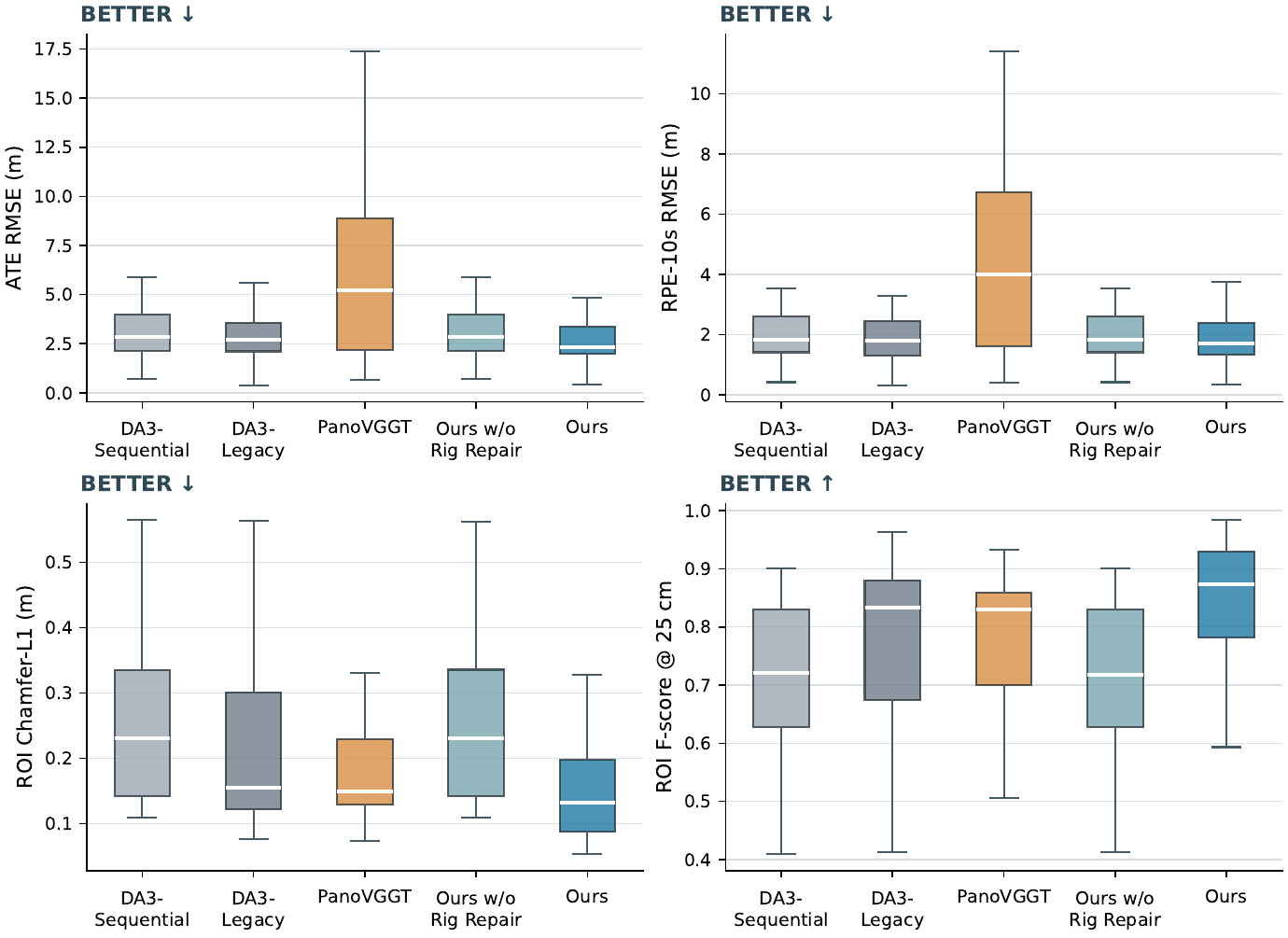}
    \caption{Per-sequence distributions of trajectory and geometry metrics.}
     \label{fig:boxplot}
\end{figure}

The quantitative results against the strongest baselines are presented in Tab.~\ref{tab:main_results}. Values are sequence medians with 95\% percentile bootstrap intervals from 10,000 run-resampled replicates (seed 2027). Paired differences use the favorable-positive convention: baseline minus ours for errors and ours minus baseline for $F$-score. Additionally, the plots in Fig.~\ref{fig:boxplot} present a more detailed overview of the results. 

Against the controlled DA3-Legacy baseline, our method improves ATE on 16/30 runs, with a paired median reduction of 0.011 m [-0.013, 0.225]. It also improves ROI CD on 26/29 runs by 0.034 m [0.014, 0.047] and F@25 on 26/29 runs by 0.039 [0.015, 0.074]. At the aggregate level, Ours achieves the best trajectory accuracy, reducing median ATE from 2.704 m to 2.312 m and RPE-10s from 1.797 m to 1.712 m relative to DA3-Legacy.

The gains are also reflected in ROI geometry. Ours obtains the lowest CD (0.132 m) and the highest precision, recall, and F@25 (0.900, 0.842, and 0.873, respectively). While PanoVGGT attains competitive geometric accuracy (0.149 m CD), its substantially larger trajectory errors indicate that strong local geometric reconstruction does not necessarily translate into accurate global motion estimation. Ours instead provides the strongest overall balance between trajectory and geometric accuracy.

The results further show that rig repair is beneficial. Removing rig repair degrades CD from 0.132 m to 0.230 m and F@25 from 0.873 to 0.718, highlighting the importance of correcting the rig geometry.

\begin{table*}[t]
  \centering
  \caption{Quantitative results. We report medians over 30 trajectories and 29 runs with LiDAR geometry. Every cloud is independently registered under the same scale-adjusting evaluation protocol. Best values within each block are bold.}
  \label{tab:main_results}
  \small
  \begin{minipage}[t]{0.42\textwidth}
    \centering
    \textbf{Trajectory}\par\vspace{2pt}
    \setlength{\tabcolsep}{3.2pt}
    \resizebox{\linewidth}{!}{%
    \begin{tabular}{lcc}
      \toprule
      Method & ATE RMSE [m] $\downarrow$ & RPE-10s RMSE [m] $\downarrow$ \\
      \midrule
      DA3-Sequential & 2.841 & 1.827 \\
      DA3-Legacy & 2.704 & 1.797 \\
      PanoVGGT & 5.206 & 3.991 \\
      \midrule
      Ours w/o Rig Repair & 2.841 & 1.827 \\
      Ours & \textbf{2.312} & \textbf{1.712} \\
      \bottomrule
    \end{tabular}%
    }
  \end{minipage}
  \hfill
  \begin{minipage}[t]{0.55\textwidth}
    \centering
    \textbf{ROI geometry}\par\vspace{2pt}
    \setlength{\tabcolsep}{3.2pt}
    \resizebox{\linewidth}{!}{%
    \begin{tabular}{lcccc}
      \toprule
      Method & CD [m] $\downarrow$ & P@25 cm $\uparrow$ & R@25 cm $\uparrow$ & F@25 cm $\uparrow$ \\
      \midrule
      DA3-Sequential & 0.231 & 0.748 & 0.711 & 0.720 \\
      DA3-Legacy & 0.155 & 0.837 & 0.802 & 0.833 \\
      PanoVGGT & 0.149 & 0.841 & 0.802 & 0.830 \\
      \midrule
      Ours w/o Rig Repair & 0.230 & 0.748 & 0.707 & 0.718 \\
      Ours & \textbf{0.132} & \textbf{0.900} & \textbf{0.842} & \textbf{0.873} \\
      \bottomrule
    \end{tabular}%
    }
  \end{minipage}
\end{table*}

\begin{figure*}[t]
     \centering
     \includegraphics[width=\textwidth]{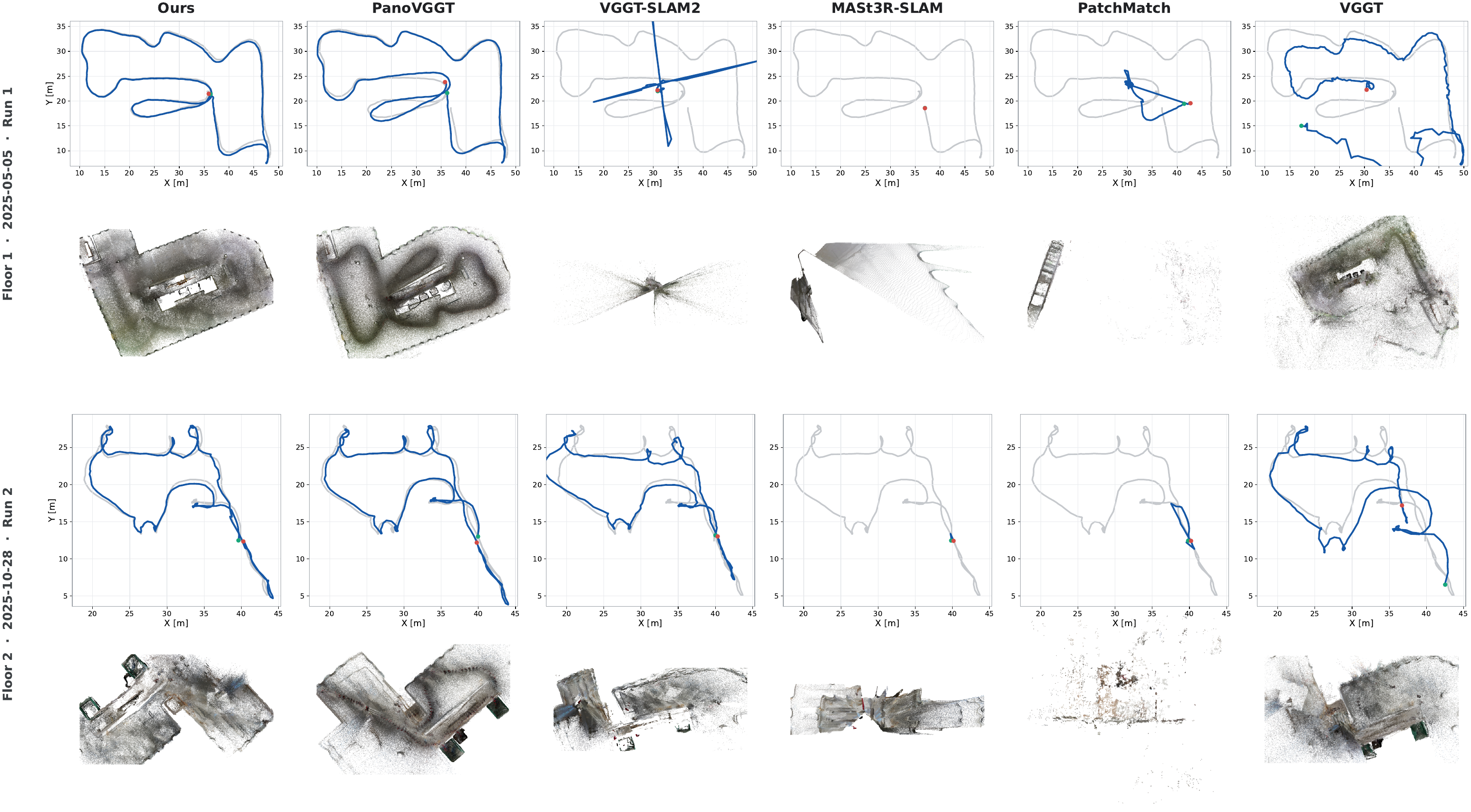}
    \caption{Qualitative comparison on two challenging sequences. For each sequence, the upper row shows the estimated trajectory and the lower row shows the reconstructed point cloud. Methods that did not consistently produce complete long-sequence reconstructions are included for qualitative analysis but omitted from the main quantitative table.}
     \label{fig:qualitative}
\end{figure*}

\paragraph{Qualitative results.}
Figure~\ref{fig:qualitative} compares the estimated trajectories and reconstructed 3D point clouds on a representative challenging sequence, selected to make differences in long-range consistency visually apparent. Among the compared baselines, PanoVGGT produced the reconstruction closest to ours and is therefore shown alongside our result. The remaining methods exhibited substantially larger failures in either trajectory estimation or reconstructed geometry on this sequence and are omitted from the quantitative comparison for clarity.

\subsection{Ablation Study}

We conduct an ablation study to quantify the contribution of the individual components of the proposed pipeline to the overall reconstruction performance. Results are presented in Tab.~\ref{tab:ablations}.

\paragraph{Capture-level cyclic retrieval.}
Using identical frozen SALAD descriptors, we evaluate capture pairs separated by at least 80 frames. GT is used only offline to label pairs within $1.5\,\mathrm{m}$ as positives and beyond $3\,\mathrm{m}$ as negatives, yielding 28 evaluable runs. As shown in Tab.~\ref{tab:ablations}, cyclic four-view scoring improves both pair AP and recall at 99\% precision, with higher AP on 24/28 runs. These metrics evaluate retrieval ranking rather than final loop-edge correctness.

\paragraph{Validated virtual-rig repair.}
To isolate repair, we compare configurations whose effective settings differ
only in whether accepted rig corrections update view poses and pointmaps
during sequential reconstruction and joint loop verification. Enabling repair
lowers median ROI CD from $0.230$ to $0.132\,\mathrm{m}$ and raises
F@25cm from $0.718$ to $0.873$. CD improves on $28/29$ runs and
F@25cm on $29/29$. These complete-cloud metrics show improved
aligned fused geometry.

\paragraph{Verified global graph update.}
We compare the final graph-optimized trajectory with its pre-graph
counterpart, which uses identical local predictions and virtual-rig repair
but does not apply the verified non-local loop constraints or global graph
optimization. Over all 30 runs, the graph update reduces median ATE from
$2.841$ to $2.312\,\mathrm{m}$ and RPE-10s from $1.827$ to
$1.712\,\mathrm{m}$. The paired reductions are $0.176\,\mathrm{m}$
[$0.000$, $0.560$] for ATE and $0.088\,\mathrm{m}$
[$-0.006$, $0.209$] for RPE-10s. Thus, the gain results from the verified
global update rather than a different local reconstruction.

\begin{table}[t]
  \centering
  \caption{Controlled component experiments. Retrieval uses 28 runs with
  GT-labeled revisits, geometry uses 29 LiDAR runs, and trajectory uses all 30
  runs. We additionally report sequence-wise win/tie/loss (W/T/L) counts for
  the \emph{first} metric in each ablation block. A win indicates that the full method
  outperforms the compared variant on that sequence according to the metric
  direction, a loss indicates the opposite, and values differing by less than
  $10^{-12}$ are counted as ties. The rig-repair block is an end-to-end
  ablation: loop verification and graph construction are recomputed after
  repair is disabled.}
  \label{tab:ablations}
  \small
  \setlength{\tabcolsep}{3.3pt}
  \begin{tabular}{lccc}
    \toprule
    Retrieval & Pair AP $\uparrow$ & R@P99 $\uparrow$ & W/T/L \\
    \midrule
    Single-view max & 0.530 & 0.178 & -- \\
    Cyclic four-view & \textbf{0.609} & \textbf{0.190} & 24/0/4 \\
    \midrule
    Rig repair & ROI CD $\downarrow$ & ROI F@25\,$\mathrm{cm}$ $\uparrow$ & W/T/L \\
    \midrule
    Off & 0.230 & 0.718 & -- \\
    On & \textbf{0.132} & \textbf{0.873} & 28/0/1 \\
    \midrule
    Graph update & ATE RMSE $\downarrow$ & RPE-10s RMSE $\downarrow$ & W/T/L \\
    \midrule
    Pre-graph & 2.841 & 1.827 & -- \\
    Ours & \textbf{2.312} & \textbf{1.712} & 20/0/10 \\
    \bottomrule
  \end{tabular}
\end{table}

\subsection{Limitations}

RIGOR relies on several assumptions that restrict its current applicability. First, the method assumes gravity-aligned panoramic input and a known virtual-rig configuration; substantial leveling errors or deviations from the assumed camera geometry may therefore reduce the reliability of rig fitting and repair. Second, although the proposed loop verification is designed to reject perceptually similar but geometrically inconsistent locations, highly repetitive environments can still lead to missed or incorrect loop candidates. The method also inherits the limitations of the frozen feed-forward reconstruction backbone, including errors in predicted depth, intrinsics, and local camera poses that cannot always be corrected using the rig constraints alone.

Finally, the current evaluation focuses on construction environments from a single benchmark and uses an evaluation-only similarity registration for geometric comparison. Consequently, the reported geometry metrics characterize reconstructed shape after global alignment rather than the recovery of absolute metric scale or map placement. Evaluating the approach across a broader range of environments and cameras, as well as improving metric-scale recovery, are important directions for future work.

\section{Conclusion}

We presented RIGOR, a rig-informed framework for improving the consistency of feed-forward 3D reconstruction from gravity-aligned omnidirectional videos. By decomposing each panorama into four perspective views with a shared optical center and known relative orientations, RIGOR introduces explicit geometric constraints without modifying or fine-tuning of the underlying reconstruction network. We exploit this virtual-rig structure to detect and repair locally inconsistent predictions, retrieve loop closures through cyclically aligned multi-view descriptors, and geometrically verify revisits before incorporating them into a $\mathrm{Sim}(3)$ pose graph whose scale variables remain free to correct accumulated scale drift. This allows reliable reconstruction in long sequences in environments poor in texture. Experiments on construction-site sequences from the Hilti-Trimble-Oxford dataset \cite{hiltichallenge} demonstrate that incorporating known panoramic rig geometry can improve both trajectory and reconstruction consistency. More broadly, our results highlight the value of combining learned feed-forward geometry with lightweight, physically grounded constraints, offering a practical path toward robust panoramic reconstruction without panorama-specific training.

{
    \small
    \bibliographystyle{ieeenat_fullname}
    \bibliography{main}

@String{CVPR={Proc. of Conference on Computer Vision and Pattern Recognition (CVPR)}}

@String{ECCV={Proc. of European Conference on Computer Vision (ECCV)}}

@String{ICLR={Proc. of International Conference on Learning Representations (ICLR)}}

@String(CVPR  = {CVPR})

@String(ECCV  = {ECCV})

@String{ICLR = {ICLR}}

@InProceedings{salad,
    author    = {Izquierdo, Sergio and Civera, Javier},
    title     = {Optimal Transport Aggregation for Visual Place Recognition},
    booktitle = {Proceedings of the IEEE/CVF Conference on Computer Vision and Pattern Recognition (CVPR)},
    month     = {June},
    year      = {2024},
}

@misc{hiltichallenge,
    title = {{Hilti}-{Trimble}-{Oxford} Dataset: 360 Visual-Inertial Benchmark with Floor Plan Priors for SLAM and Localization},
    author = {Centanni, Samuele and Zhang, Yuhao and Tao, Yifu and Kindle, Julien and Neuhaus, Frank and Koß, Tilman and Patel, Aryaman and Helmberger, Michael and Szymańska, Emilia and Gräber, Torben and Fallon, Maurice},
    year = {2026},
    eprint = {2607.06464},
    url = {https://arxiv.org/abs/2607.06464}
}

@inproceedings{panovggt,
  author    = {Guo, Yijing and Chao, Mengjun and Wang, Luo and
               Zhao, Tianyang and Dai, Haizhao and Zhang, Yingliang and
               Yu, Jingyi and Shi, Yujiao},
  title     = {{PanoVGGT}: Feed-Forward {3D} Reconstruction from Panoramic Imagery},
  booktitle = {Proceedings of the IEEE/CVF Conference on Computer Vision and Pattern Recognition (CVPR)},
  year      = {2026},
  pages     = {36444--36453}
}

@inproceedings{vggt,
  author    = {Wang, Jianyuan and Chen, Minghao and Karaev, Nikita and
               Vedaldi, Andrea and Rupprecht, Christian and Novotny, David},
  title     = {{VGGT}: Visual Geometry Grounded Transformer},
  booktitle = {Proceedings of the IEEE/CVF Conference on Computer Vision and Pattern Recognition (CVPR)},
  year      = {2025},
  pages     = {5294--5306}
}

@inproceedings{da3,
  author    = {Lin, Haotong and Chen, Sili and Liew, Jun Hao and
               Chen, Donny Y. and Li, Zhenyu and Zhao, Yang and
               Peng, Sida and Guo, Hengkai and Zhou, Xiaowei and
               Shi, Guang and Feng, Jiashi and Kang, Bingyi},
  title     = {Depth Anything 3: Recovering the Visual Space from Any Views},
  booktitle = {International Conference on Learning Representations (ICLR)},
  year      = {2026}
}

@inproceedings{mapanything,
  author    = {Keetha, Nikhil and M{\"u}ller, Norman and
               Sch{\"o}nberger, Johannes and Porzi, Lorenzo and
               Zhang, Yuchen and Fischer, Tobias and Knapitsch, Arno and
               Zauss, Duncan and Weber, Ethan and Antunes, Nelson and
               Luiten, Jonathon and Lopez-Antequera, Manuel and
               Rota Bul{\`o}, Samuel and Richardt, Christian and
               Ramanan, Deva and Scherer, Sebastian and Kontschieder, Peter},
  title     = {{MapAnything}: Universal Feed-Forward Metric {3D} Reconstruction},
  booktitle = {International Conference on 3D Vision (3DV)},
  year      = {2026},
  pages     = {499--509},
  organization = {IEEE},
  doi       = {10.1109/3DV69130.2026.00054}
}

@inproceedings{pi3,
  author    = {Wang, Yifan and Zhou, Jianjun and Zhu, Haoyi and
               Chang, Wenzheng and Zhou, Yang and Li, Zizun and
               Chen, Junyi and Pang, Jiangmiao and Shen, Chunhua and
               He, Tong},
  title     = {{$\pi^3$: Permutation-Equivariant Visual Geometry Learning}},
  booktitle = {International Conference on Learning Representations (ICLR)},
  year      = {2026}
}

@inproceedings{vggt360,
  author    = {Yuan, Jiayi and Jiang, Haobo and Soh, De Wen and Zhao, Na},
  title     = {{VGGT-360}: Geometry-Consistent Zero-Shot Panoramic Depth Estimation},
  booktitle = {Proceedings of the IEEE/CVF Conference on Computer Vision and Pattern Recognition (CVPR)},
  year      = {2026},
  pages     = {19874--19883}
}

@inproceedings{mast3rslam,
  author    = {Murai, Riku and Dexheimer, Eric and Davison, Andrew J.},
  title     = {{MASt3R-SLAM}: Real-Time Dense {SLAM} with {3D} Reconstruction Priors},
  booktitle = {Proceedings of the IEEE/CVF Conference on Computer Vision and Pattern Recognition (CVPR)},
  year      = {2025},
  pages     = {16695--16705}
}

@article{vggtslam2,
  author  = {Maggio, Dominic and Carlone, Luca},
  title   = {{VGGT-SLAM 2.0}: Real-time Dense Feed-forward Scene Reconstruction},
  journal = {Robotics: Science and Systems},
  year    = {2026}
}

@misc{panoair,
  author        = {Wu, Yiyang and Zhang, Xiaohu and Du, Yanjin and
                   Zhang, Tongsu and Li, Chujun and Chen, Siyang and
                   Zhang, Guoyi and Xu, Xiangpeng},
  title         = {{PanoAir}: A Panoramic Visual-Inertial {SLAM} with Cross-Time Real-World {UAV} Dataset},
  year          = {2026},
  eprint        = {2604.00852},
  archivePrefix = {arXiv},
  primaryClass  = {cs.RO}
}

@misc{survey360,
      title={Deep Learning for Omnidirectional Vision: A Survey and New Perspectives}, 
      author={Hao Ai and Zidong Cao and Jinjing Zhu and Haotian Bai and Yucheng Chen and Lin Wang},
      year={2022},
      eprint={2205.10468},
      archivePrefix={arXiv},
      primaryClass={cs.CV},
      url={https://arxiv.org/abs/2205.10468}, 
}

@misc{picard2023surveyrealtime3dscene,
      title={A survey on real-time 3D scene reconstruction with SLAM methods in embedded systems}, 
      author={Quentin Picard and Stephane Chevobbe and Mehdi Darouich and Jean-Yves Didier},
      year={2023},
      eprint={2309.05349},
      archivePrefix={arXiv},
      primaryClass={cs.RO},
      url={https://arxiv.org/abs/2309.05349}, 
}

@Article{survey3d,
AUTHOR = {Liu, Shuai and Yang, Mengmeng and Xing, Tingyan and Yang, Ran},
TITLE = {A Survey of 3D Reconstruction: The Evolution from Multi-View Geometry to NeRF and 3DGS},
JOURNAL = {Sensors},
VOLUME = {25},
YEAR = {2025},
NUMBER = {18},
ARTICLE-NUMBER = {5748},
ISSN = {1424-8220},
DOI = {10.3390/s25185748}
}

@inproceedings{dust3r,
  author = {Wang, Shuzhe and Leroy, Vincent and Cabon, Yohann and Chidlovskii, Boris and Revaud, Jerome},
  title = {{DUSt3R}: Geometric {3D} Vision Made Easy},
  booktitle = {Proceedings of the IEEE/CVF Conference on Computer Vision and Pattern Recognition (CVPR)},
  year = {2024},
  pages = {20697--20709},
  doi = {10.1109/CVPR52733.2024.01956}
}

@inproceedings{mast3r,
  author    = {Leroy, Vincent and Cabon, Yohann and Revaud, Jerome},
  title     = {Grounding Image Matching in {3D} with {MASt3R}},
  booktitle = {Proceedings of the European Conference on Computer Vision (ECCV)},
  year      = {2024},
  pages     = {71--91},
  doi       = {10.1007/978-3-031-73220-1_5}
}

@inproceedings{must3r,
  author = {Cabon, Yohann and Stoffl, Lucas and Antsfeld, Leonid and Csurka, Gabriela and Chidlovskii, Boris and Revaud, Jerome and Leroy, Vincent},
  title = {MUSt3R: Multi-view Network for Stereo 3D Reconstruction},
  booktitle = {Proceedings of the IEEE/CVF Conference on Computer Vision and Pattern Recognition (CVPR)},
  year = {2025},
  pages = {1050--1060},
  doi = {10.1109/CVPR52734.2025.00106}
}

@inproceedings{spann3r,
  author = {Wang, Hengyi and Agapito, Lourdes},
  title = {3D Reconstruction with Spatial Memory},
  booktitle = {Proceedings of the International Conference on 3D Vision (3DV)},
  year = {2025},
  pages = {78--89}
}

@inproceedings{cut3r,
  author = {Wang, Qianqian and Zhang, Yifei and Holynski, Aleksander and Efros, Alexei A. and Kanazawa, Angjoo},
  title = {Continuous 3D Perception Model with Persistent State},
  booktitle = {Proceedings of the IEEE/CVF Conference on Computer Vision and Pattern Recognition (CVPR)},
  year = {2025},
  pages = {10510--10522}
}

@inproceedings{slam3r,
  author = {Liu, Yuzheng and Dong, Siyan and Wang, Shuzhe and Yin, Yingda and Yang, Yanchao and Fan, Qingnan and Chen, Baoquan},
  title = {SLAM3R: Real-Time Dense Scene Reconstruction from Monocular RGB Videos},
  booktitle = {Proceedings of the IEEE/CVF Conference on Computer Vision and Pattern Recognition (CVPR)},
  year = {2025},
  pages = {16651--16662},
  doi = {10.1109/CVPR52734.2025.01552}
}

@inproceedings{megasam,
  author = {Li, Zhengqi and Tucker, Richard and Cole, Forrester and Wang, Qianqian and Jin, Linyi and Ye, Vickie and Kanazawa, Angjoo and Holynski, Aleksander and Snavely, Noah},
  title = {MegaSaM: Accurate, Fast, and Robust Structure and Motion from Casual Dynamic Videos},
  booktitle = {Proceedings of the IEEE/CVF Conference on Computer Vision and Pattern Recognition (CVPR)},
  year = {2025},
  pages = {10486--10496}
}

@inproceedings{openvslam,
  author = {Sumikura, Shinya and Shibuya, Mikiya and Sakurada, Ken},
  title = {OpenVSLAM: A Versatile Visual SLAM Framework},
  booktitle = {Proceedings of the 27th ACM International Conference on Multimedia},
  series = {MM '19},
  year = {2019},
  pages = {2292--2295},
  publisher = {ACM},
  address = {New York, NY, USA},
  doi = {10.1145/3343031.3350539}
}

@article{omnids0,
  author = {Matsuki, Hidenobu and von Stumberg, Lukas and Usenko, Vladyslav and Stueckler, Joerg and Cremers, Daniel},
  title = {Omnidirectional DSO: Direct Sparse Odometry With Fisheye Cameras},
  journal = {IEEE Robotics and Automation Letters},
  volume = {3},
  number = {4},
  year = {2018},
  pages = {3693--3700},
  doi = {10.1109/LRA.2018.2855443}
}

@INPROCEEDINGS{patchmatch,
  author={Surmann, Hartmut and Thurow, Marc and Slomma, Dominik},
  booktitle={2022 IEEE International Symposium on Safety, Security, and Rescue Robotics (SSRR)},
  title={PatchMatch-Stereo-Panorama, a fast dense reconstruction from 360° video images},
  year={2022},
  volume={},
  number={},
  pages={366-372},
  doi={10.1109/SSRR56537.2022.10018698}}

@misc{ren2024grounded,
      title={Grounded SAM: Assembling Open-World Models for Diverse Visual Tasks}, 
      author={Tianhe Ren and Shilong Liu and Ailing Zeng and Jing Lin and Kunchang Li and He Cao and Jiayu Chen and Xinyu Huang and Yukang Chen and Feng Yan and Zhaoyang Zeng and Hao Zhang and Feng Li and Jie Yang and Hongyang Li and Qing Jiang and Lei Zhang},
      year={2024},
      eprint={2401.14159},
      archivePrefix={arXiv},
      primaryClass={cs.CV}
}
}

\clearpage
\setcounter{page}{1}
\maketitlesupplementary
\setcounter{section}{0}
\renewcommand{\thesection}{\Alph{section}}

\section{Parameter Selection Strategy}
\label{sec:supp-selection-v7}

This supplement explains two operating choices used by RIGOR: the perspective
field of view and the sequential block layout.  Within each comparison, the
input samples and frozen
DA3 setup are kept fixed while the factor under consideration is varied.
These studies document the rationale for two operating choices used
uniformly across the evaluation sequences.  We focus on the observed design
trade-offs and leave the method and benchmark results to the main paper.

\section{Choosing the Perspective Field of View}
\label{sec:supp-fov-selection-v7}

\paragraph{Design question.}
Wider perspective crops create more shared content between adjacent yaw views,
but may also move farther from the perspective-image regime in which the
frozen predictor is most reliable.  We therefore ask how much overlap can be
added before four-view pose consistency begins to deteriorate.

\paragraph{Controlled comparison.}
We compare horizontal FoVs of $90^\circ$, $95^\circ$, $100^\circ$,
$105^\circ$, and $110^\circ$. The render size ($768\!\times\!512$), sampled
panoramas, and frozen DA3 setup are fixed. Pose consistency is evaluated on the same
100 panoramas, with 25 sampled from each of four sequences used for this
diagnostic. A common test compares the raw DA3 poses with the known four-view
rig using a $0.1$ m center criterion and a $5^\circ$ rotation criterion.

As a permissive image-space proxy, we evaluate the four cyclically adjacent
yaw-view pairs for each panorama, giving 400 pairs per FoV. We call a pair
supported if the homography estimator returns a non-empty consensus set; the
supported-pair fraction is the fraction of these 400 pairs that are supported.
This is not a measure of verified physical overlap: at $90^\circ$, adjacent
crops only meet at their boundaries, and non-empty responses can arise from
boundary sampling or ambiguous matches in repetitive scenes.

\begin{figure*}[t]
  \centering
  \includegraphics[width=0.96\textwidth]{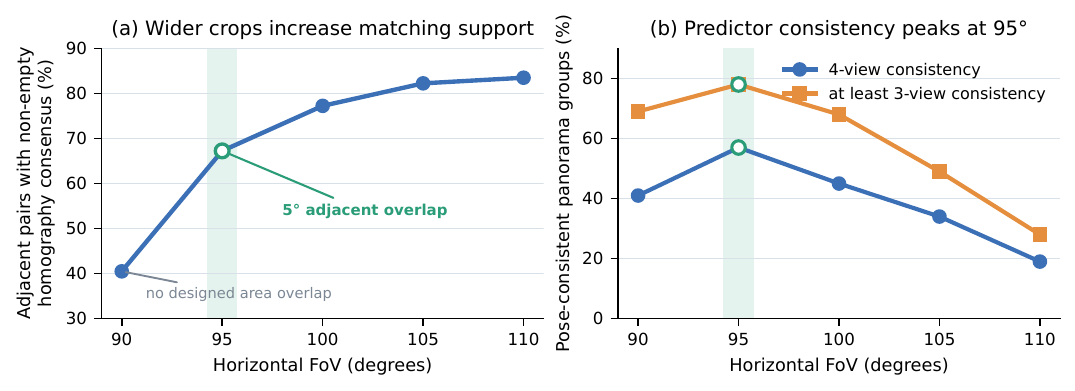}
  \vspace{0.05cm}
  \includegraphics[width=0.50\textwidth]{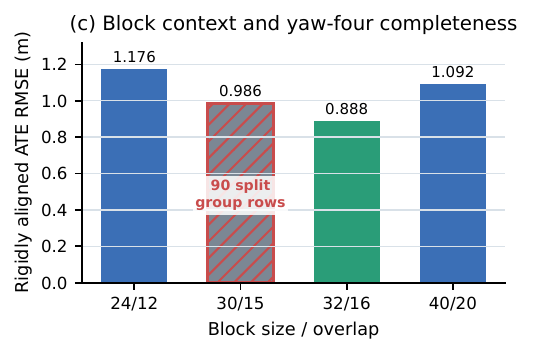}
  \caption{Parameter choices. (a) The fraction of 400 cyclically adjacent-view
  pairs per FoV returning a non-empty homography consensus increases with
  wider crops. At $90^\circ$, the crops have no designed area overlap; non-empty
  responses can arise from boundary sampling or ambiguous matches.  (b) Raw
  DA3 pose consistency on the same 100 panoramas peaks at $95^\circ$ under the
  common $0.1$ m center and $5^\circ$ rotation diagnostic.  These rates describe
  the parameter comparison, not accepted repairs.  (c) On one 229-panorama run
  under the earlier sequential/no-loop protocol, yaw-four-aligned layouts avoid
  incomplete capture groups, and 32/16 gives the lowest ATE in this check.}
  \label{fig:supp-parameters-v7}
\end{figure*}

\paragraph{Observation.}
Figure~\ref{fig:supp-parameters-v7}(a) shows that image-space support increases
with FoV: the supported-pair fraction rises from $40.5\%$ at $90^\circ$ to
$67.3\%$ at $95^\circ$ and $83.5\%$ at $110^\circ$. In contrast,
Fig.~\ref{fig:supp-parameters-v7}(b) shows a non-monotonic pose-consistency
trend. Four-view / at-least-three-view consistency changes from $41\%/69\%$
at $90^\circ$ to $57\%/78\%$ at $95^\circ$, then falls to $45\%/68\%$ at
$100^\circ$ and $19\%/28\%$ at $110^\circ$.

\paragraph{Chosen setting.}
We use $95^\circ$: it introduces designed overlap between adjacent crops and
gives the highest four-view and at-least-three-view consistency rates in the
tested sweep, before wider crops begin to degrade the predicted rig geometry.

\section{Choosing Block Size and Overlap}
\label{sec:supp-block-selection-v7}

\paragraph{Design question.}
A sequential block should provide temporal context without splitting the four
perspective views that represent one panorama.  We ask which block layout
preserves that unit while maintaining stable sequential reconstruction.

\paragraph{Controlled comparison.}
We compare block/overlap settings 24/12, 30/15, 32/16, and 40/20 on the same
229-panorama sequence (916 perspective views), using $95^\circ$ rendering and
a sequential reconstruction with loop closure disabled.  The input views and
all reconstruction settings other than block size and overlap are identical.

\paragraph{Observation.}
The 30/15 layout produces 90 incomplete yaw-four group rows at block
boundaries.  The three layouts whose block and overlap sizes are multiples of
four produce none.  As a supporting trajectory check on this run, their
rigidly aligned ATE RMSE values are 1.176, 0.986, 0.888, and 1.092 m for
24/12, 30/15, 32/16, and 40/20, respectively.

\paragraph{Chosen setting.}
We use 32/16 because it preserves every four-view capture and provides 50\%
overlap.  Figure~\ref{fig:supp-parameters-v7}(c) shows that it also gives the
lowest rigidly aligned ATE RMSE in this single-sequence check; preserving
complete four-view captures remains the primary consideration.

\end{document}